\documentclass[10pt,twocolumn,letterpaper]{article}

 \usepackage[pagenumbers]{iccv} 
\usepackage{array}
\usepackage{makecell}
\usepackage{algorithm}
\usepackage{algpseudocode}
\usepackage{graphicx}
\usepackage{multirow}
\usepackage{wrapfig}
\definecolor{iccvblue}{rgb}{0.21,0.49,0.74}
\usepackage[pagebackref,breaklinks,colorlinks,allcolors=iccvblue]{hyperref}

\def\paperID{14469} 
\def\confName{ICCV}
\def\confYear{2025}

\def\method{SQIL\xspace}

\title{Saliency-Aware Quantized Imitation Learning for Efficient Robotic Control}

\author{Seongmin Park$^{1}$,    Hyungmin Kim$^{1}$, Sangwoo Kim$^{1}$,    Wonseok Jeon$^{2}$,    Juyoung Yang$^{2}$, \\ {Byeongwook Jeon$^{2}$},    {Yoonseon Oh$^{1}$} and {Jungwook Choi$^{1}$}\thanks{Corresponding author}  \\
        \normalsize{\textsuperscript{1}Hanyang University},
        \normalsize{\textsuperscript{2}Hyundai Motor Company} \\
        \normalsize{Seoul, Republic of Korea}\\
        \small{\textsuperscript{1}\texttt{\{skstjdals, kong4274, kimz1121, yoh21\}@hanyang.ac.kr}} \\
        \small{\textsuperscript{2}\texttt{\{wsjeon, yjy6711, smiler\}@hyundai.com}, \textsuperscript{1*}\texttt{choij@hanyang.ac.kr}} \\   
}

\begin{document}
\maketitle
\begin{abstract}

Deep neural network (DNN)-based policy models, such as vision-language-action (VLA) models, excel at automating complex decision-making from multi-modal inputs. However, scaling these models greatly increases computational overhead, complicating deployment in resource-constrained settings like robot manipulation and autonomous driving. To address this, we propose Saliency-Aware Quantized Imitation Learning (\method), which combines quantization-aware training with a selective loss-weighting strategy for mission-critical states. By identifying these states via saliency scores and emphasizing them in the training loss, \method preserves decision fidelity under low-bit precision. We validate \method's generalization capability across extensive simulation benchmarks with environment variations, real-world tasks, and cross-domain tasks (self-driving, physics simulation), consistently recovering full-precision performance. Notably, a 4-bit weight-quantized VLA model for robotic manipulation achieves up to 2.5$\times$ speedup and 2.5$\times$ energy savings on an edge GPU with minimal accuracy loss. These results underline \method’s potential for efficiently deploying large IL-based policy models on resource-limited devices.

\end{abstract} 
\section{Introduction}
\label{sec:introduction}

In recent years, deep neural network (DNN)-based policy models have significantly advanced robot manipulation and autonomous driving~\cite{zhang2021end,kim24openvla,driess2023palm,brohan2022rt,brohan2023rt,liang2018cirl}, primarily by surpassing traditional search-based methods through imitation learning (IL) from expert data. This success has led to growing interest in adopting foundation models for large-scale IL, aiming to improve generalization and robustness beyond limited data, facilitating policy transfer across different robot embodiments, tasks, or environments~\cite{brohan2022rt, brohan2023rt}. Vision-Language Action (VLA) models~\cite{kim24openvla,driess2023palm,brohan2022rt,brohan2023rt} extend pre-trained vision-language models to robotics via next-token prediction, integrating visual and textual information for enhanced manipulation capabilities. Despite their potential as foundation models supporting cross-embodiment transfer, IL-based VLA models often suffer from slow inference speeds, high computational costs, and substantial memory usage~\cite{wen2024tinyvla}, making deployment on resource-constrained, battery-powered robots challenging.

Quantization techniques reduce DNN inference costs by converting full-precision (FP)\footnote{We use BFloat16~\cite{kalamkar2019study} and Float32 as baseline FP for OpenVLA and CILRS, unless noted otherwise.} weights and activations to lower precision~\cite{choi2018pact,esser2020learned,lin2023awq,frantar2023optq}, and extensive research focuses on mitigating the resulting numerical errors to preserve accuracy. Post-training quantization (PTQ) adjusts weights and activations to reduce quantization loss, whereas quantization-aware training (QAT) incorporates quantization directly in training to enhance robustness. These strategies are well studied in computer vision~\cite{yuan2022ptq4vit,nagel2020up,li2021brecq,zhou2024lidar} and natural language processing~\cite{lin2023awq,frantar2022gptq,xiao2023smoothquant}, but their effects on IL-based applications, such as robot manipulation or autonomous driving, remain underexplored.

\begin{figure*}[t]
\centering
{\includegraphics[width=2\columnwidth]{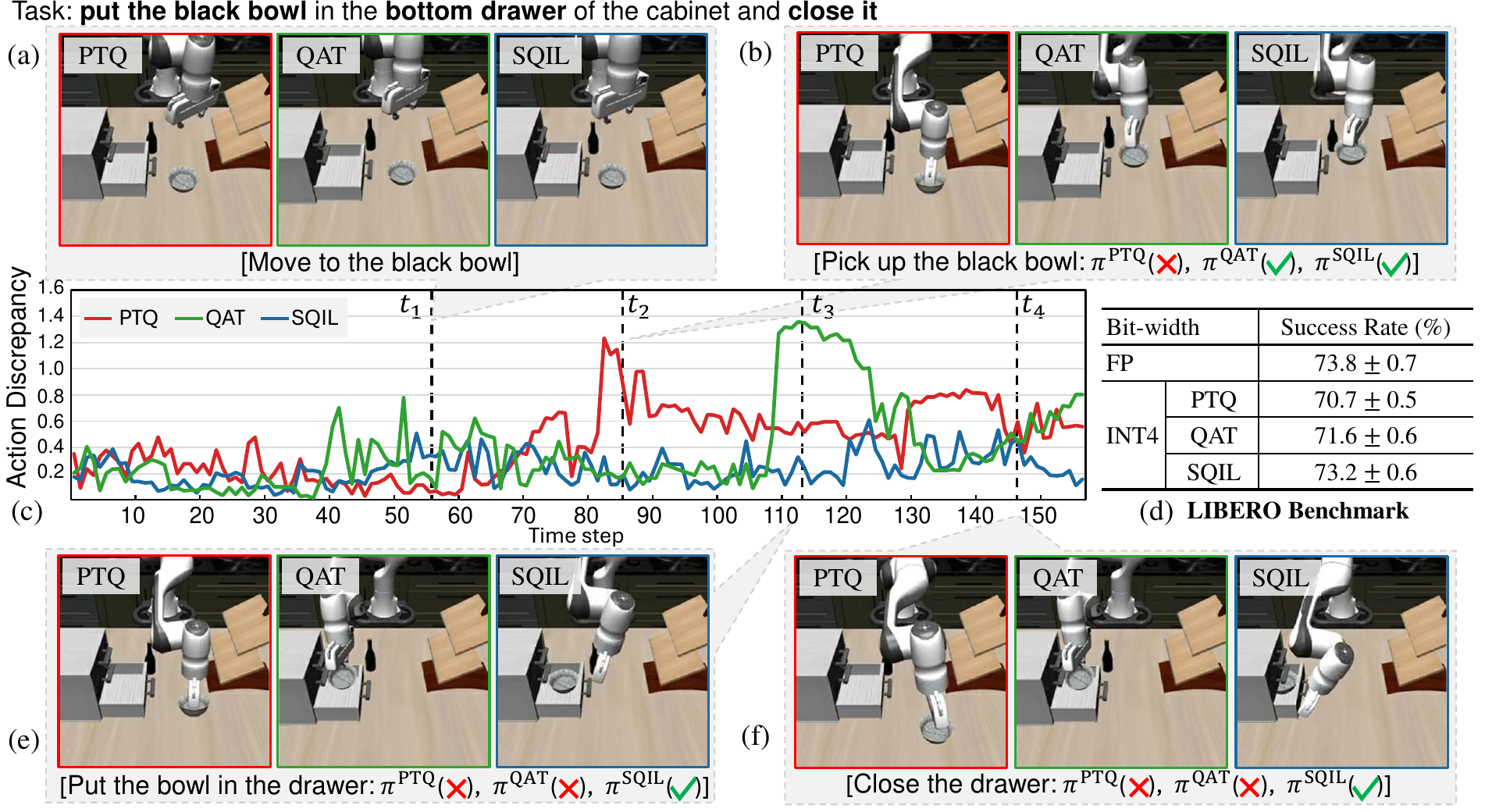}}
\vspace{-0.2cm}
\caption{Comparison of action discrepancy (L2-norm) between the quantized and full-precision (FP) policies in \textit{OpenVLA} using the PTQ, QAT, and \method, and evaluation of the average success rate on the LIBERO benchmark. For fair comparison, episode timelines are segmented into four sub-tasks and realigned based on successful transitions.}
\label{fig:bench_difficulty_comparision}

\end{figure*}

In this work, we investigate how quantization errors affect IL policy decisions and note two key observations: 1) Quantization errors generally have minor impact across most timesteps, producing small action discrepancies (e.g., $t_1$ in Fig.\ref{fig:bench_difficulty_comparision}). 2) Certain critical states, however, experience large deviations in actions due to quantization errors (Fig.\ref{fig:bench_difficulty_comparision}-$t_2$: PTQ, $t_3$: QAT), ultimately causing mission failures (Fig.\ref{fig:bench_difficulty_comparision}-$t_4$). This differs from typical long-sequence prediction, where errors accumulate at each timestep to deviate from the FP policy’s action distribution; here, the quantized policy must specifically address a small number of mission-critical states.

To address this, we present saliency-aware quantized imitation learning (\method), a simple yet effective method for IL-based robotic control. \method integrates quantization into fine-tuning, increasing robustness against quantization errors. Unlike traditional QAT for supervised learning, \method incorporates quantization-robust action distillation (QRD): 1) identifying mission-critical states via saliency scores, and 2) applying an importance-weighted loss to emphasize correct actions at these states. Consequently, the quantized policy more closely follows the FP policy’s decisions for crucial states. As seen in Fig.\ref{fig:bench_difficulty_comparision}, \method suppresses action discrepancies at critical timesteps ($t_2$, $t_3$), leading to mission success and fully recovering the FP’s success rate (Fig.\ref{fig:bench_difficulty_comparision}(d)).

We validate \method's generalization capability across extensive simulation benchmarks with environment variations, real-world tasks, and cross-domain tasks (self-driving, physics simulation), consistently recovering full-precision performance. For robot manipulation, evaluations of \textit{OpenVLA}\cite{kim24openvla} on the LIBERO benchmark\cite{liu2024libero} show that our 4-bit weight-quantized models achieve success rates comparable to the FP policy, delivering a 2.5× speedup and 2.5× energy savings on edge GPUs. In autonomous driving, tests of \textit{CILRS}\cite{zhang2021end} on the NoCrash benchmark\cite{codevilla2019exploring} confirm that our 4-bit weight- and activation-quantized policy maintains the FP policy's performance, realizing a 3.7× speedup and 3.1× energy savings on low-end GPUs. These findings mark the first successful demonstration of recovering and efficiently deploying quantized IL-based policies.

\noindent\textbf{Contributions.}
Building on the observations and results above, this paper makes four primary contributions:
\begin{itemize}
    \item \textbf{First Systematic Study of Quantized IL.}  
    To our knowledge, this is the \textit{first comprehensive analysis of quantized IL} that identifies the significance of \textit{mission-critical states}—moments where fine-grain control is essential. We found that most failures in quantized policies stem from coarse action control during physical interactions (e.g., grasping or releasing objects). This insight explains \textit{why naive quantization fails}.
    \item \textbf{SIS: Policy-Driven Critical State Detection.}
    SIS uses the policy’s action sensitivity to surface mission-critical states, surpassing vision-language key-frame (KF) detectors~\cite{kou2024kisa}. 
    \item \textbf{SQIL: Saliency-Aware Quantized Imitation Learning.}  
    By coupling 4-bit QAT with SIS-weighted loss (QRD), SQIL achieve a 2–4$\times$ speedup and energy savings  while maintaining a success rate within 1\% of the FP baseline. 
    \item \textbf{Broad, Cross-Domain Validation.} Experiments on robot manipulation, autonomous driving, and MuJoCo control confirm SQIL’s generality in both simulation and real-world.  
\end{itemize}


\section{Related Works}
\label{sec:related}

\begin{figure*}[t]

{\includegraphics[width=2.05\columnwidth]{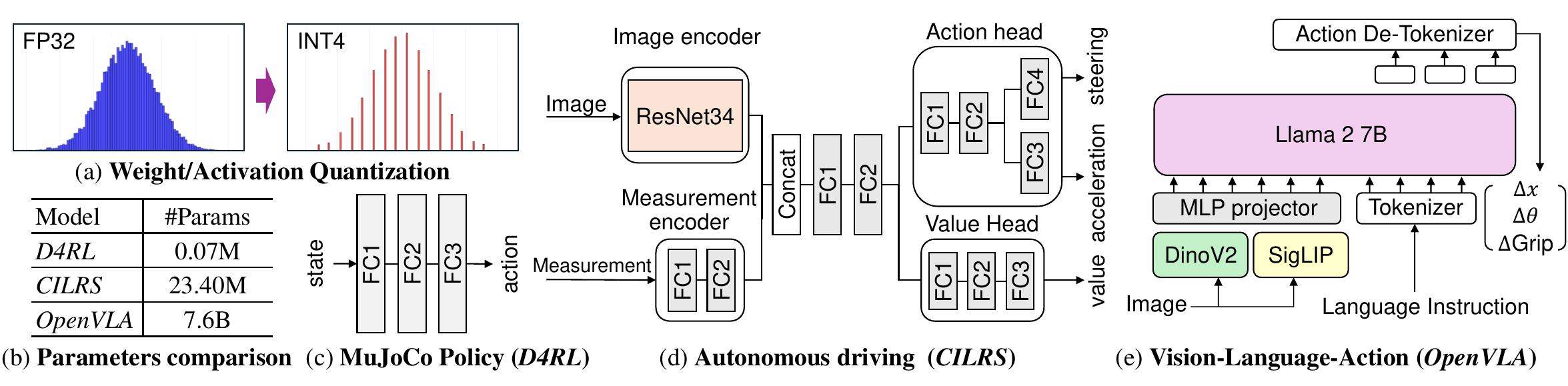}}
\vspace{-0.3 cm}
\caption{Comparison of the structure, number of parameters of DNN-based policy models.}
\label{fig:d4rl_ad_vla}
\vspace{-0.3 cm}
\end{figure*}

\subsection{DNN-based Policy for Imitation Learning}
\label{sec:dnn-based_policy}
DNN-based policy models trained via imitation learning (IL) have propelled significant progress in physics simulation, autonomous driving, and robotic manipulation. In MuJoCo~\cite{todorov2012mujoco}, for example, DNN models control robots in continuous joint spaces by processing state-vector inputs, often trained on benchmarks such as \textit{D4RL}\cite{fu2020d4rl}. Autonomous driving models like \textit{CILRS}\cite{zhang2021end} learn to predict acceleration and steering from camera images and measurement vectors, focusing on safe driving. In robotic manipulation, vision-language-action models such as RT-2~\cite{brohan2023rt} and \textit{OpenVLA}\cite{kim24openvla} use LLMs to fuse visual and textual inputs for multi-task tasks; \textit{OpenVLA} specifically generates actions (x, y, z, roll, pitch, yaw, grasp) auto-regressively from image observations and language instructions. These IL-based models learn by imitating expert behaviors from pre-collected datasets. For instance, \textit{CILRS} is trained on 80 demonstration episodes, while \textit{OpenVLA} is pretrained on 970k robot episodes and then fine-tuned on 1,700 LIBERO simulation episodes~\cite{liu2024libero} with LoRA~\cite{hu2021lora} for training efficiency.

As IL applications grow more complex, DNN-based policy models have dramatically increased in size. Fig.\ref{fig:d4rl_ad_vla}(c-e) shows representative IL models for physics simulation (D4RL), autonomous driving (\textit{CILRS}), and vision-language-action (\textit{OpenVLA}), with model sizes in Fig.\ref{fig:d4rl_ad_vla}(b). D4RL relies on a three-layer fully connected network, while \textit{CILRS} combines a ResNet-based image encoder with fully connected layers for steering control. By contrast, \textit{OpenVLA} merges the Llama2-7B language model~\cite{touvron2023llama2} with visual features from DINOv2~\cite{oquab2023dinov2} and SigLIP~\cite{zhai2023sigmoid}, creating a versatile foundation for robotic manipulation. This expansion has increased model sizes by about 100,000 times (from 0.07M parameters in D4RL to 7.6B in \textit{OpenVLA}). Since scaling model size often boosts capability~\cite{kaplan2020scaling}, there is a critical need for efficient inference methods that reduce computational and memory loads on resource-constrained devices.

\begin{figure*}[t!]
    \centering
    \includegraphics[width=0.85\linewidth]{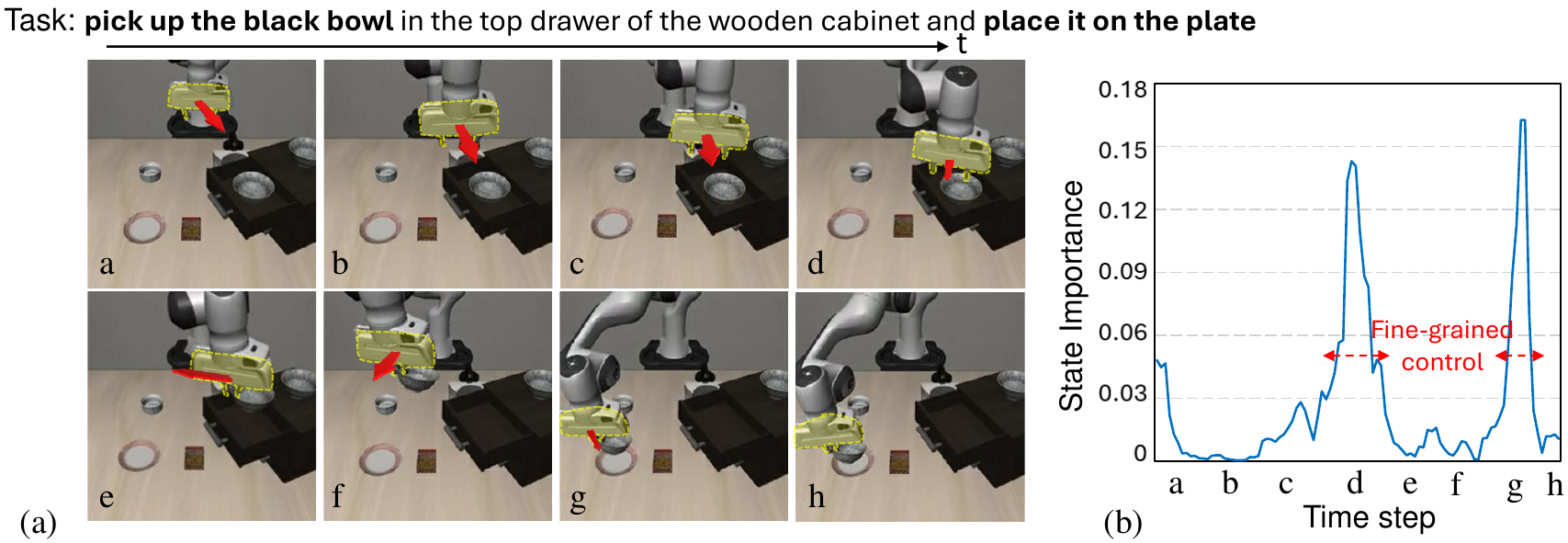}
\vspace{-0.2cm}
   \caption{Comparison of timestep-wise state importance in the \textit{OpenVLA} robot manipulation task.}
\vspace{-0.3 cm}
\label{fig:STATE_importance}
\end{figure*}

\subsection{Quantization for Efficient Policy Inference}
Quantization reduces neural networks' computational and memory requirements by lowering the bit-precision of weights and/or activations, typically applied to the input operands of general matrix multiplication (GEMM). In compute-intensive convolutional models, both weights and activations are quantized~\cite{he2016deep,chollet2017xception,simonyan2014very}, increasing operations per second~\cite{markidis2018nvidia} and boosting energy efficiency~\cite{horowitz20141}. Large language models (LLMs), on the other hand, often adopt weight-only quantization~\cite{lin2023awq,frantar2023optq} to reduce memory-access bottlenecks. 

Numerical errors can degrade accuracy since quantization maps continuous values to discrete states (Fig.\ref{fig:d4rl_ad_vla}(a)). Post-training quantization (PTQ) employs data manipulations~\cite{krishnamoorthi2018quantizing,banner2019post,tambe2020algorithm}—for instance, AWQ~\cite{lin2023awq} shapes weight distributions for weight-only quantization in LLMs—though it does not inherently improve resilience to quantization errors\footnote{We use AWQ~\cite{lin2023awq} and RTN~\cite{jacob2018quantization} as our baseline PTQ methods for OpenVLA and others, respectively, unless otherwise noted.}. In contrast, quantization-aware training (QAT) incorporates quantization effects during training to maintain accuracy~\cite{hwang2014fixed,zhou2016dorefa,choi2018pact,esser2020learned}, sometimes combined with knowledge distillation (KD)~\cite{mckinstry2018low,zhuang2018towards,zhao2023poster} for further error compensation. For LLMs, where tuning all parameters is expensive, LoRA~\cite{dettmers2024qlora,li2023loftq,gunter2024apple} refines only small adapters while keeping base parameters in reduced precision.

Despite rising demand for efficient policy inference, most quantization approaches remain tailored to reinforcement learning (RL), which relies on direct environment interaction—unavailable in IL—to mitigate quantization errors. Early work like low-precision policy distillation (LPPD)\cite{mckinstry2018low} used aggressive 1-bit QAT for classical policy models. More recent RL studies\cite{faust2022quarl,zhang2023fastact} have quantized lightweight policies, leveraging environment interaction to compensate for errors; structured pruning~\cite{park2024pruning} similarly reduces model size without hurting decision quality. However, these methods do not directly apply to IL, which lacks environment feedback. OpenVLA~\cite{kim24openvla} explored QAT with LoRA for 4-bit quantization but saw mixed success. Our investigation shows that conventional QAT suffers noticeable accuracy loss on OpenVLA, 
underscoring the need for more advanced quantization-error mitigation methods for IL.

\section{Background and Challenges}
\label{sec:background}

\subsection{Imitation Learning}

Imitation Learning (IL) enables an agent to learn policies directly from expert demonstrations, which is particularly useful when reward functions are hard to define (e.g., robotics, autonomous driving). Let \(S\) be the set of states, and \(A\) the set of actions. The expert dataset \(D_E = \{\tau_1, \ldots, \tau_N\}\) contains \(N\) expert demonstrations, each \(\tau_i\) being a sequence of state-action pairs of length \(T_i\), denoted as \(\tau_i = \{(s_1, a_1), \ldots, (s_{T_i}, a_{T_i})\}\), where \(s_t \in S\) and \(a_t \in A\). 

The goal of IL is to learn a policy \(\pi^{\text{FP}}_\theta: S \to A\) that replicates expert behavior. A common approach is \textit{behavior cloning}~\cite{osa2018algorithmic}, which maximizes the likelihood of expert state-action pairs via:
\begin{equation}
\label{eq:il_objective}
\mathcal{L}^{\text{IL}}(\theta) = - \mathbb{E}_{\tau_i \sim \mathcal{D}_{\text{E}}} \left[ \frac{1}{|\tau_i|} \sum_{(s_t,a_t) \in \tau_i} \log \pi^{\text{FP}}_{\theta}(a_t|s_t)\right].
\end{equation}
Minimizing this loss guides \(\pi^{\text{FP}}_\theta\) to imitate the expert actions \(a_t\) for corresponding states \(s_t\).

\subsection{Policy Quantization}

As IL extends to more complex robotic control tasks, DNN-based policy models can reach billions of parameters, demanding substantial computation and memory. To address this, quantization is applied to weights and/or activations (e.g., weight and activation quantization for compute-bounded CNNs, weight-only quantization for memory-bounded LLMs). We convert full-precision data \( w^{FP} \) to \( b \)-bit precision using:
\begin{equation}
\label{eq:quantization}
w^{Q} = \text{Clip}\left(\left\lfloor \frac{w^{FP}}{\gamma} \right\rceil, -2^{b-1}, 2^{b-1}-1\right),
\end{equation}
where \(\gamma\) is the quantization scale. The resulting quantized policy model $\pi^{\text{Q}}_\theta$ reduces computation and memory use but may produce action discrepancies due to quantization errors. QAT incorporates quantization into IL with:
\begin{equation}
\label{eq:qat_objective}
\mathcal{L}^{\text{QAT}}(\theta) = - \mathbb{E}_{\tau_i \sim \mathcal{D}_{\text{E}}} \left[ \frac{1}{|\tau_i|} \sum_{(s_t,a_t) \in \tau_i} \log \pi^{\text{Q}}_{\theta}(a_t|s_t)\right].
\end{equation}
This allows \(\pi^{\text{Q}}\) to approximate expert behavior similar to \(\pi^{\text{FP}}\). The same expert dataset $\mathcal{D}_E$ is reused to update $\pi^{\text{Q}}_{\theta}$'s weights, and the same hyperparameters (learning rate, etc.) can typically be used without hassle.

\subsection{Challenges}

Although PTQ and QAT mitigate some quantization errors, they still face noticeable performance drops, especially in success rate. As shown in Fig.~\ref{fig:bench_difficulty_comparision}, comparing the action distributions of a full-precision VLA policy with INT4 PTQ and QAT reveals two main points: 1) quantization errors usually result in minor discrepancies across most timesteps ($t_1$); and 2) at certain \textit{mission-critical} states, such as ``Pick-up the black bowl'' ($t_2$) or ``Put the bowl in the drawer'' ($t_3$),  quantization causes large action deviations ($\pi^{\text{PTQ}}$ at $t_2$ and $\pi^{\text{QAT}}$ at $t_3$), leading to mission failures  ($t_4$)\footnote{A similar effect is observed in self-driving tasks.}.

These observations differ from \textit{distribution shift}~\cite{kumar2019stabilizing}, which arises when encountering unseen or out-of-distribution states. Here, the deviation stems from model quantization rather than dataset mismatch (QAT still uses the same expert dataset), and only a few critical states fail instead of the entire trajectory. These observations suggest quantized policies must specifically address those critical states. Developing methods that minimize quantization errors at these points is crucial for reliable robotic control performance in dynamic, complex environments.

\section{Method}
\label{sec:method}

Motivated by the previous observations, we propose enabling the quantized policy to learn compensations for quantization errors in mission-critical states. The proposed method, saliency-aware quantized imitation learning (\method), consists of two key components: 1) a saliency-based state-importance score (SIS) to identify mission-critical states without supervision, and 2) a quantization-robust action distillation (QRD) method that selectively emphasizes learning from the SIS-identified mission-critical states.

\begin{figure}[t]

\centering
{\includegraphics[width=1\columnwidth]{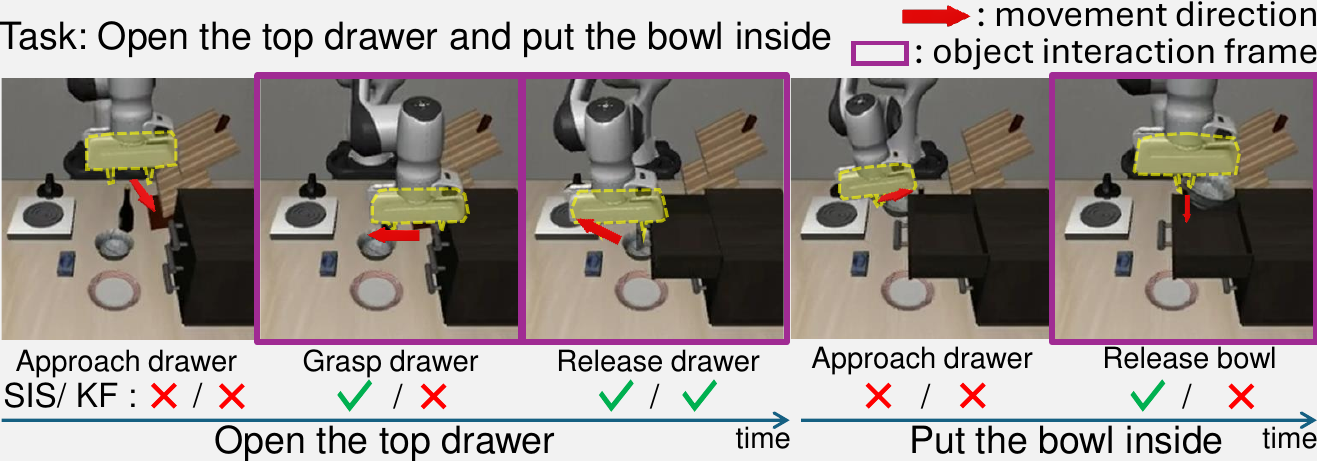}}

\caption{Comparison of SIS and keyframe (KF) on trajectory.}

\label{fig:sis_key}
\end{figure}  
\begin{table}[t]
\centering
\resizebox{0.95\linewidth}{!}{
\begin{tabular}{c|ccc}
\hline
LIBERO Benchmark & QAT & SQIL (KF) & \textbf{SQIL (SIS)} \\ \hline
Avg. Success Rate (\%) & 71.6 ± 0.5 & 72.6 ± 0.7 & \textbf{73.2 ± 0.6} \\
\hline
\end{tabular}}
\caption{Saliency metric comparison: Keyframe (KF) and SIS.}
\label{tab:sis_vs_key}
\end{table}

\subsection{Saliency-based State-Importance Score}

From Fig.~\ref{fig:bench_difficulty_comparision}, we see that mission-critical states depend on both the task and the environment, making it non-trivial to identify which states need extra attention during imitation learning. Therefore, we quantify the state importance score (SIS) to identify mission-critical states. Motivated by \cite{greydanus2018visualizing}, we maintain that states showing large action discrepancies under visual perturbation are mission-critical, as these discrepancies reveal the crucial visual regions for successful decision-making. We measure the perturbation-based action saliency at each position of a state as follows:
\begin{equation}
\label{eq:saliency-score}
S_{\pi}(s_t,k) = \frac{1}{2} \left\|\pi(s_t) - \pi\left(\phi(s_t,k)\right)\right\|^2,
\end{equation}
where \(\phi(s_t, k)\) applies a perturbation (e.g. a Gaussian blur at position \(k\). Higher \(S_{\pi}(s_t,k)\) indicates that local modifications at \(k\) have a greater impact on the policy's output 

We then define SIS as the average saliency across all positions \(k\):
\begin{equation}
\label{eq:state-importance}
SIS^{s_t}_{\pi} = \mathbb{E}_{k} \left[ S_{\pi^\text{FP}}(s_t,k) \right].
\end{equation}
A higher SIS suggests that the state has more perturbation-sensitive positions, often found in mission-critical scenarios. As shown in Fig.~\ref{fig:STATE_importance}, SIS effectively highlights the states at the ``Pick-Up'' (timestep $d$) and ``Drop'' (timestep $g$) actions, highlighting its ability to detect moments requiring fine-grained control.
Compared to vision-language-based key-frame (KF) detection~\cite{kou2024kisa}, which often activates only at coarse sub-task boundaries (e.g., “drawer open”, “object placed”), SIS responds to subtle yet decisive robot–environment interactions—such as initial grasp or release—by measuring how small perturbations alter the policy’s output (Fig.\ref{fig:sis_key}). Quantitatively, replacing SIS with KF in \method\ reduces success by 1.1\% on LIBERO manipulation tasks (Table\ref{tab:sis_vs_key}), confirming that policy-driven saliency is essential to robust performance under quantization.

\begin{algorithm}[t]
\caption{Saliency-Aware Quantized Imitation Learning}
\begin{algorithmic}
\State \textbf{Input:} Pre-trained policy $\pi^{\text{FP}}$, quantized policy $\pi^{\text{Q}}_{\theta}$, Expert dataset $\mathcal{D}_{\text{E}}$, Number of epochs $N$
\State \textbf{Output:} Quantized policy $\pi^{\text{Q}}_{\theta}$

\State Initialize $\pi^{\text{Q}}_{\theta}$ from $\pi^{\text{FP}}$
\State Calculate $SIS_{\pi^{\text{FP}}}$ with $\pi^{\text{FP}}$, $\mathcal{D}_{\text{E}}$ using Eq.~(\ref{eq:state-importance})
            
\For{epoch = 1 to $N$}
    \For{each trajectory $\tau_i$ in $\mathcal{D}_{\text{E}}$}
        \For{each state-action pair $(s_t, a_t)$ in $\tau_i$}
            \State Calculate $\mathcal{L}^{\text{QAT}}$ using Eq.~(\ref{eq:qat_objective})
            \State Calculate $\mathcal{L}^{\text{QRD}}$ with $SIS_{\pi^{FP}}$ using Eq.~(\ref{eq:QRD_objective})
            \State $\mathcal{L}^{\text{\method}}=\mathcal{L}^{\text{QAT}}+\mathcal{L}^{\text{QRD}}$ (Eq.~(\ref{eq:total_objective}))
            \State Compute $\frac{\partial \mathcal{L}^{\text{\method}}}{\partial \theta} \approx \frac{\partial \mathcal{L}^{\text{\method}}}{\partial \theta_{\text{Q}}}$
            \State Update $\pi^{\text{Q}}_{\theta}$ with $\frac{\partial \mathcal{L}^{\text{\method}}}{\partial \theta}$
        \EndFor
    \EndFor
\EndFor

\State \textbf{return} Quantized policy $\pi^{\text{Q}}_{\theta}$

\end{algorithmic}
\label{algo:alg2}
\end{algorithm}

\subsection{Quantization-Robust Action Distillation}
\label{sec:QRD}

We selectively focus on mission-critical states using the state importance score (SIS). To achieve this, we propose quantization-robust action distillation (QRD), which leverages the FP policy and demonstration data to reduce quantization errors by distilling the FP policy’s action distributions into the quantized model. Its loss function measures the discrepancy between the quantized and FP policies’ action distributions:
\begin{align}
\label{eq:QRD_objective}
\mathcal{L}^{\text{QRD}}(\theta) = \alpha_t \cdot \mathbb{E}_{\tau_i \sim \mathcal{D}_{\text{E}}} \left[ \frac{1}{|\tau_i|} \sum_{s_t \in \tau_i}D({\pi}^{\text{Q}}(s_t), {\pi}^{\text{FP}}(s_t)) \right], 
\end{align}
where $\alpha_t = \beta$ if $SIS^{s_t}_{\pi^\text{FP}} > T$, and $\alpha_t = 1$ otherwise; 
$D$ is a discrepancy metric (e.g., L2-norm), and ${\pi}(s_t)$ is the probability distribution of all possible actions for state $s_t$. The hyperparameter \(\beta (> 1)\) applies extra weight to high-importance states, and \(T\) is a threshold that selects the top $p$=20\% of SIS values. Unlike conventional KD methods, QRD employs the weighting $\alpha_t$ to amplify the loss for states identified by SIS (i.e., $SIS^{s_t}_{\pi^\text{FP}} > T$). Training convergence is not sensitive to the hyperparameters $D(),\beta,T$, and we use the same values (details in Supplementary Sec.~\ref{appendix:ablation}) for all the experiments across robot control, self-driving, and physics simulation.

\begin{figure*}
    \centering
    \includegraphics[width=0.9\linewidth]{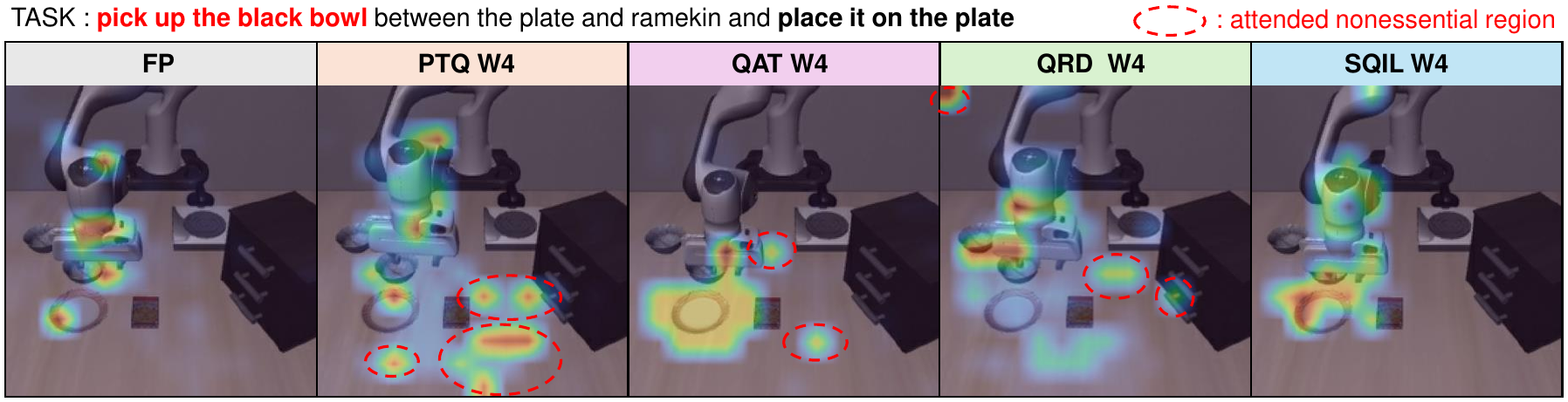}
\vspace{-0.1 cm}
   \caption{Comparison of attention visualization in \textbf{important states} (pick up the black bowl) for tasks successfully completed on the LIBERO-Spatial benchmark. Additional examples are provided in the \textit{supplementary materials.}}
\vspace{-0.2 cm}
\label{fig:attention}
\end{figure*}

\subsection{Saliency-Aware Quantized Imitation Learning}
To enhance conventional QAT for mission-critical states, we introduce saliency-aware quantized imitation learning (\method), which selectively strengthens IL's weight adjustment for salient states. \method's loss function combines QAT and QRD:
\begin{equation}
\label{eq:total_objective}
\mathcal{L}^{\text{\method}}(\theta)  = \mathcal{L}^{\text{QAT}}(\theta)+\mathcal{L}^{\text{QRD}}(\theta),
\end{equation} 
\setlength{\columnsep}{15pt}
This combined approach effectively reduces quantization errors, enabling the quantized policy to generalize comparably to the FP policy. $\mathcal{L}^{\text{QAT}}$ maximizes the quantized policy's log-likelihood of expert actions, while $\mathcal{L}^{\text{QRD}}$ aligns the quantized policy’s action distribution with the FP policy. By applying selective weighting $\alpha_t$, QRD emphasizes states requiring precise control instead of uniformly minimizing the discrepancy across all states. 

Algorithm~\ref{algo:alg2} outlines the overall \method procedure. We reuse the same expert dataset and training hyperparameters, and SIS can be precomputed once for \(\mathcal{D}_E\) and $\pi^\text{FP}$. Consequently, \method offers a turnkey solution for efficiently quantizing an FP policy using an existing expert dataset. As shown in Fig.\ref{fig:bench_difficulty_comparision}, \method suppresses action discrepancies at mission-critical timesteps ($t_2$, $t_3$), leading to mission success and fully recovering the FP’s success rate (Fig.\ref{fig:bench_difficulty_comparision}(d)).

\subsection{Analysis}
\label{sec:analysis}

In this section, we quantitatively answer two questions about \method's efficacy: 1) Does it restore the disrupted saliency of the quantized policy? 2) Do QAT and QRD within \method work synergistically to recover the action density of the quantized policy?

\textbf{Saliency Visualization}.
To qualitatively assess how quantization affects IL policy behavior and how \method addresses this issue, we visualize the saliency map using Eq.~\ref{eq:saliency-score}. Fig.~\ref{fig:attention} shows a robot executing ``Pick-up the black bowl'' under different policies. The FP policy exhibits high saliency on objects of interest (e.g., the robot arm, the target bowl, and the destination plate). In contrast, the PTQ policy focuses on irrelevant regions, indicating that quantization errors misidentify salient areas. Although QAT and QRD alone attempt to restore expert or FP actions, applying loss with uniform importance across all states limits their ability to fix the saliency. In contrast, \method's saliency map closely mirrors the FP policy, indicating that it effectively recovers the quantized policy’s desired focus. We confirm this quantitatively in Table~\ref{tab:attsim}, which shows \method achieving lower average divergence from the FP saliency map than PTQ.

\begin{table}
\centering
\resizebox{0.85\columnwidth}{!}{%
\begin{tabular}{l|c|c|c|c}
\Xhline{2\arrayrulewidth}
\multirow{3}{*}{Method} & \multicolumn{4}{c}{Average Saliency Divergence$\downarrow$} \\ \cline{2-5} 
& \makecell{LIBERO\\-Spatial} & \makecell{LIBERO\\-Object} & \makecell{LIBERO\\-Goal} & \makecell{LIBERO\\-Long}\\ \midrule
PTQ & 0.0821 & 0.0268 & 0.0612 &  0.0640\\ \midrule
\method & 0.0559& 0.0198 & 0.0458  &  0.0521\\ \Xhline{2\arrayrulewidth}
\end{tabular}
}
\vspace{-0.2 cm}
\caption{Comparison of AttDiv with \textit{OpenVLA} on the LIBERO.}
\vspace{-0.3 cm}
\label{tab:attsim}
\end{table}
\begin{figure}[t]
\centering
{\includegraphics[width=0.75\columnwidth]{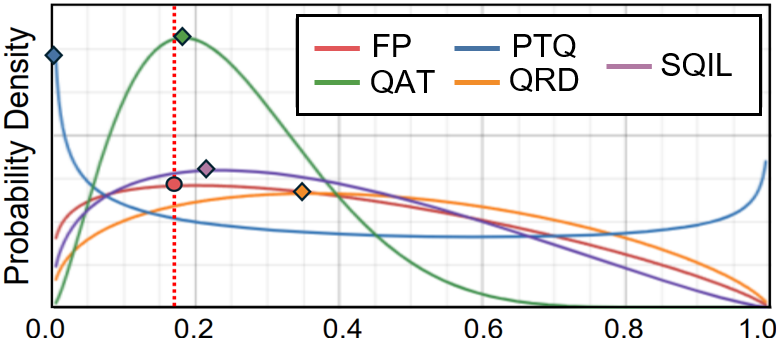}}
\caption{Comparison of action distributions for various quantization methods using \textit{CILRS} (W4A4) on NoCrash-dense at state \(s_j\). The red circle denotes action \(a_j\) from the demonstration dataset $\mathcal{D}_{\text{E}}$, and the diamond shape marks the highest probability point for each distribution.}
\vspace{-0.2 cm}
\label{fig:act_dist}
\end{figure}
\begin{table*}
\centering
\resizebox{1.55\columnwidth}{!}{%
\begin{tabular}{l|c|c|c|c|c}
\Xhline{2\arrayrulewidth}
\multirow{2}{*}{Method} & \multirow{2}{*}{Quantizer (INT4)} & \multicolumn{4}{c}{Success Rate \% $\uparrow$ } \\ \cline{3-6} 
 &  & LIBIERO-Spatial  & LIBERO-Object & LIBERO-Goal & LIBERO-Long  \\ \midrule
FP & - & 84.0 ± 0.9 & 83.9 ± 0.3 & 76.6  ± 0.6  & 50.7  ± 1.2   \\ \midrule
PTQ & AWQ  & 80.1 ± 0.5  & 81.3 ± 0.4  & 74.3 ± 0.6  & 47.2 ± 0.6    \\ 
QAT & AWQ  & 80.9 ± 0.7  & 82.4 ± 0.4  & 75.7 ± 0.5  & 47.3 ± 0.3   \\ 
\method & AWQ  & \textbf{83.9  ± 0.5}  & \textbf{83.5  ± 0.}5  & \textbf{76.3  ± 0.4}  & \textbf{49.2 ± 1.0 }  \\ \midrule
PTQ & QuaRot  & 81.2  ± 1.0  & 81.7  ± 0.6  & 74.8  ± 0.4  & 47.6  ± 0.9    \\
QAT & QuaRot  & 81.8  ± 0.9  & 82.8  ± 0.5  & 75.2  ± 0.6  & 48.0  ± 0.8    \\
\method & QuaRot  & \textbf{83.8  ± 0.8}  & \textbf{83.7  ± 0.5}  & \textbf{76.3  ± 0.5}  &  \textbf{49.4  ± 0.9}    \\ \Xhline{2\arrayrulewidth}
\end{tabular}
}
\caption{Comparison of success rate across various INT4 quantization with \textit{OpenVLA} and scenarios in the LIBERO benchmark.}
\label{tab:libero_comparison}
\end{table*}

\begin{table*}[t]
\centering
\resizebox{1\linewidth}{!}{%
\begin{tabular}{l|c|c|c||l|c|c}
\Xhline{2\arrayrulewidth}
\multicolumn{4}{c||}{Language Instruction Generalization} & \multicolumn{3}{c}{Illumination Generalization} \\ 
\Xhline{2\arrayrulewidth}
\multirow{2}{*}{Task} & \multirow{2}{*}{\#Trials} & \multicolumn{2}{c||}{\# Successes $\uparrow$} & Light & \multicolumn{2}{c}{Success Rate(\%) $\uparrow$} \\ \cline{3-4}\cline{6-7}
& &FP &\method{} W4&Intensity &FP &\method{} W4 \\ \Xhline{2\arrayrulewidth}
\textbf{Pick up} the black bowl on the wooden cabinet and \textbf{place it} on the plate & 50 & 34 & 29 & 100\% & 50.7 & 49.2 \\
\textbf{Grab} the black bowl from the wooden cabinet and \textbf{set it} on the plate & 50 & 25 & 28 & 80\% & 50.5 & 49.1 \\
\textbf{Remove} the black bowl from the wooden cabinet and \textbf{place it} on the plate & 50 & 31 & 34 & 60\% & 49.5 & 48.8 \\
\textbf{Lift} the black bowl from the wooden cabinet and \textbf{set it} on the plate & 50 & 28 & 27 & -- & -- & -- \\
\Xhline{2\arrayrulewidth}
\end{tabular}
}
\caption{Generalization comparisons of \textit{OpenVLA}: Language instruction generalization in LIBERO-Spatial (left) and illumination generalization in LIBERO-Long (right).}
\label{tab:Generalization2}
\vspace{-0.3cm}
\end{table*}


\textbf{Action Distribution Comparison.} 
To evaluate the synergy of QAT and QRD, we compare the action distributions of the FP policy and policies quantized by PTQ, QAT, QRD, and \method in the self-driving model (CILRS). Fig.~\ref{fig:act_dist} depicts each policy’s distribution, with circle and diamond markers indicating the maximum-likelihood actions of the FP and quantized policies, respectively. PTQ introduces large quantization errors, causing serious divergence from FP. Although QAT aligns the maximum likelihood to expert actions, it can generate overly sharp peaks that distort the FP distribution. QRD aligns the quantized distribution with the FP policy’s shape but may overlook expert actions when the FP distribution is broad. By combining both, \method preserves the FP policy’s overall behavior while emphasizing high-quality expert actions, resulting in better decisions and improved performance.

\section{Experiments}
\label{sec:experiments}
\subsection{Experimental Settings}

We conducted experiments on the quantization of DNN-based policy models across various domains, including robot manipulation, autonomous driving, and classical physics simulation. Detailed experimental settings for each task are provided in the \textit{supplementary materials}.

\textbf{Robot Manipulation:} We utilized \textit{OpenVLA}~\cite{kim24openvla} with weight quantization through AWQ~\cite{lin2023awq} and QuaRot~\cite{ashkboos2024quarot} methods. Evaluation were made using the the LIBERO~\cite{liu2024libero} benchmark in simulated environments, spanning tasks like LIBERO-Spatial, LIBERO-Object, LIBERO-Goal, and LIBERO-Long. Each task was repeated in three rounds of evaluations. We fine-tuned a minimal set of trainable parameters (110M parameters) using \method and QLoRA (r=32) and tested the model in two real-world scenarios: a replication of BridgeDataV2~\cite{walke2023bridgedata} and a tabletop setup that enables broader manipulation behaviors. A dataset $\mathcal{D}_{\text{E}}$ of 80 episodes was collected and used to further refine \textit{OpenVLA} with \method and QLoRA for deployment on the UR5 robot.


\textbf{Autonomous Driving:} We employed \textit{CILRS}~\cite{zhang2021end} and enhanced latency by applying both weight and activation quantizations at 4-bit precision, using RTN~\cite{jacob2018quantization} for PTQ and LSQ~\cite{esser2020learned} for \method. The models were evaluated on the NoCrash-dense benchmark~\cite{codevilla2019exploring}, which tests performance across different towns and weather conditions.

\textbf{Classical Physics Simulation:} Fine-tuning and assessment occurred on the D4RL benchmark~\cite{fu2020d4rl} using LSQ for both weights and activations at 4-bit precision.

\subsection{Experimentals: Robot Manipulation}

\subsubsection{Full-Benchmark Evaluation}

Table~\ref{tab:libero_comparison} outlines the comparative performance results on the LIBERO benchmark using various quantization methods. When PTQ (AWQ) applied, there is a notable decline of 3.2\% in performance relative to the baseline, a critical issue for robot manipulation tasks due to their reliance on safe real-world interactions. Conversely, QAT achieves a modest performance increase of 0.7\%, yet it still falls short of the baseline. Employing \method, which integrates additional insights from the FP policy on critical-state, significantly reduces the performance disparity to the FP model. This method shows a performance improvement of 2.3\% over AWQ, underscoring its efficacy in resource-limited settings. Furthermore, we expand our assessment to include the state-of-the-art quantizer, QuaRot~\cite{ashkboos2024quarot}, with \method consistently matching the FP baseline success rate, thus confirming its general effectiveness.

 \begin{figure}[t]
\centering
{\includegraphics[width=0.95\columnwidth]{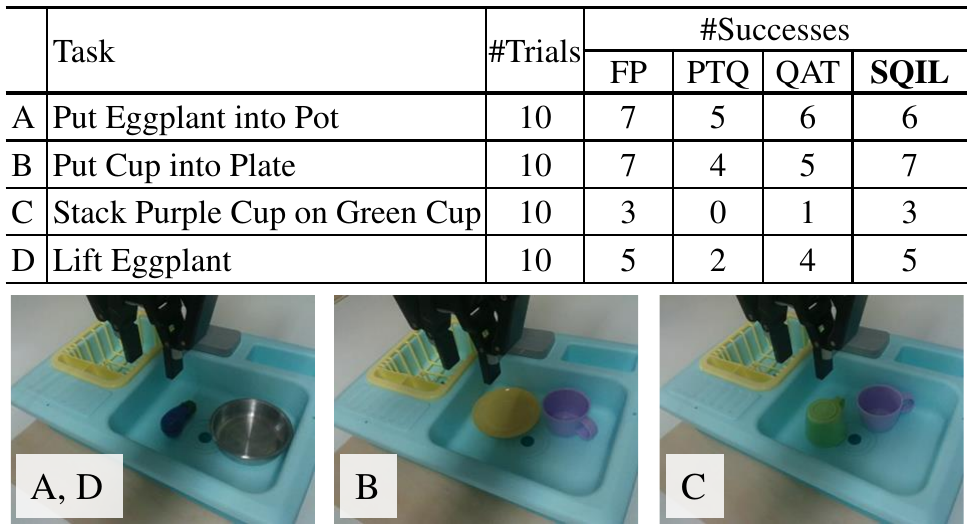}}
\vspace{-0.2 cm}
\caption{\textit{OpenVLA} real-world UR5 robot evaluation tasks and results with various INT4 quantization.}
\vspace{-0.2 cm}
\label{fig:real_world}
\end{figure}

\begin{figure}[t]
\centering
\resizebox{0.7\linewidth}{!}{%
\begin{tabular}{c|ccccc}
\Xhline{2\arrayrulewidth}
\multirow{2}{*}{Task} & \multicolumn{1}{c|}{\multirow{2}{*}{\#Trials}} & \multicolumn{4}{c}{\#Successes} \\ \cline{3-6}
                      & \multicolumn{1}{c|}{} & FP & PTQ & QAT & SQIL \\ \hline
(a) & \multicolumn{1}{c|}{30} & 25 & 22 & 25 & 24 \\ 
(b) & \multicolumn{1}{c|}{30} & 24 & 20 & 22 & 24 \\ 
(c) & \multicolumn{1}{c|}{30} & 22 & 18 & 17 & 21 \\ 
\Xhline{2\arrayrulewidth}
\multicolumn{2}{c|}{Suc.Rate(\%)} & 79 & 67 & 71 & 77 \\ 
\Xhline{2\arrayrulewidth}
\end{tabular}}
\includegraphics[width=0.9\linewidth]{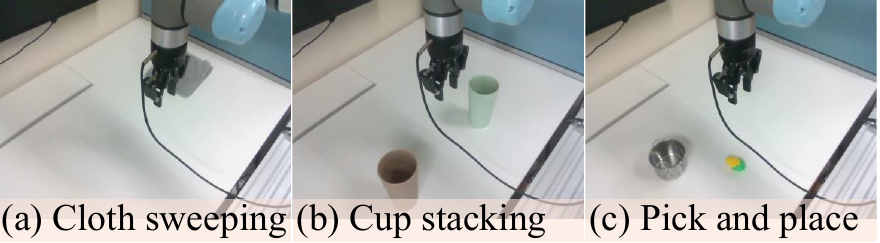}
\vspace{-0.2 cm}
\caption{Real-world evaluation with \textit{OpenVLA} on UR5.}
\vspace{-0.3 cm}
\label{fig:realworld}
\end{figure}

\begin{table}[t]

\centering
\resizebox{0.93\linewidth}{!}{
\begin{tabular}{l|c|ccc}
\hline
LIBERO Benchmark & FP & PTQ & QAT &SQIL \\ \hline
Model Size                & 3.3B & 1.1B & 1.1B & 1.1B \\
Avg. Success Rate & 93.8\% & 92.0\% & 92.6\% & \textbf{93.3\%} \\
\hline
\end{tabular}}
\caption{INT4 quantization results for $\pi_0$ with \textit{OpenVLA}.}
\label{tab:pi_0}
\vspace{-0.3cm}
\end{table}

\subsubsection{Generalization}

To assess the generalization capabilities of the quantized policy, we conduct a detailed evaluation as outlined in Table~\ref{tab:Generalization2}. We investigate the performance of the IN4 policy, quantized using \method, in responding to a variety of language instructions aimed at achieving a single goal. The results show that the IN4 policy's performance closely matches that of the FP policy. Additionally, we analyze how this policy performs under different lighting conditions, demonstrating its robustness by maintaining generalization performance at 80\% and 60\% brightness levels.

To further assess the generality of \method, we applied it to $\pi_0$~\cite{black2024pi_0}, a compact and high-performance policy obtained via flow matching-based action generation. As shown in Table~\ref{tab:pi_0}, \method consistently outperforms PTQ and QAT, achieving performance close to the FP baseline while providing 3.0$\times$ memory savings.

Fig.~\ref{fig:real_world} and Fig.~\ref{fig:realworld} compares the performance results of various quantization techniques in real-world scenarios. Relative to the FP policy, the performance of PTQ significantly declines, while QAT exhibits slight improvements. Notably, \method achieves performance comparable to the FP policy, demonstrating that reducing quantization errors in important states directly contributes to achieving goals in real-world environments.

\begin{table}[t]
\centering
\resizebox{0.95\columnwidth}{!}{%
\begin{tabular}{l|c|cccc|cccc}
\Xhline{2\arrayrulewidth}
\multirow{2}{*}{Method} & \multirow{2}{*}{Bit-width} & \multicolumn{4}{c|}{Suc. Rate \% $\uparrow$} & \multicolumn{4}{c}{Dri. Score \% $\uparrow$}  \\ \cline{3-10} 
 &  & tt & tn & nt & nn & tt & tn & nt & nn  \\ \Xhline{2\arrayrulewidth}
Baseline & FP & 82 & 74 & 80 & 68 & 78 & 71 & 80 & 64 \\ \midrule
PTQ & W4A4 & 34 & 43 & 36  & 29  & 30  & 38 & 35  & 31  \\
QAT & W4A4 & 62 & 58 & 58 & 48 & 62 & 62 & 58 & 58  \\ 
\method& W4A4 & \textbf{80 }& \textbf{72} & \textbf{72} & \textbf{68} & \textbf{76} & \textbf{70} & \textbf{74} & \textbf{61}  \\ \Xhline{2\arrayrulewidth}
\end{tabular}%
}
\vspace{-0.2 cm}
\caption{Comparative analysis of success rate, driving score, and reward across various quantization methods using \textit{CILRS} on the NoCrash-dense benchmark, differentiated by combinations of training and new conditions in town and weather (tt, tn, nt, nn).}
\vspace{-0.2 cm}
\label{tab:cilrs_main}
\end{table}


\begin{table}[t]
\centering
\resizebox{0.72\columnwidth}{!}{%
\begin{tabular}{l|c||c|c}
\Xhline{2\arrayrulewidth}
\multirow{2}{*}{TASK} & FP32 & \multicolumn{2}{c}{INT4} \\ \cline{2-4} 
 & IL & LPPD\cite{mckinstry2018low} & \method \\ \hline
Cartpole Balance & 652 & 424 & \textbf{635} \\ \hline
Walker Stand & 692 & 442 & \textbf{688} \\ \hline
Hopper Stand & 645 & 404 & \textbf{643} \\ \hline
Cheetah Run &  567 & 294 & \textbf{556} \\ \hline
Finger Spin & 684 & 421 & \textbf{640} \\ \hline
Humanoid Stand & 565 & 356 & \textbf{590} \\ \hline
Humanoid Walk & 550 & 205 & \textbf{535} \\ \Xhline{2\arrayrulewidth}
\end{tabular}%
}
\vspace{-0.2 cm}
\caption{Comparison of average return for each task by quantization method on physics simulation on MuJuCo.}
\vspace{-0.2 cm}
\label{tab:mujoco}
\end{table}
\subsection{Cross-Domain Results}
\textbf{End-to-End Autonomous Driving.}
As seen in Table~\ref{tab:cilrs_main}, the performance comparison after applying quantization methods shows a significant drop in performance in both train and new-towns under the W4A4 setting. However, by applying \method, the alignment with the driving capabilities of the FP policy is significantly improved, achieving much better performance. An example illustrating the differences in driving behavior across quantization methods is provided as a video in the \textit{supplementary materials}.

\textbf{Physics Simulation Tasks.}
To evaluate the generalization capability of our algorithms, we expand our evaluation to classical continuous control tasks from the DeepMind Control Suite within D4RL~\cite{fu2020d4rl}, as shown in Table~\ref{tab:mujoco}. This allowed us to compare our quantization techniques \method, against established methods like LPPD~\cite{mckinstry2018low}.

\subsection{Implementation}

We evaluate the versatility of our quantization techniques by implementing them on various resource-constrained platforms, including an edge device (NVIDIA Jetson AGX Orin) and a low-end GPU (NVIDIA 2080Ti). For practical deployment, we test the autonomous driving models: the W8A8 model on a CPU (ARM Cortex-A78AE) and the W4A4 model on the NVIDIA 2080Ti GPU. Additionally, 8-bit and 4-bit weight-quantized VLA models are implemented on an Jetson AGX Orin's GPU. Latency is measured using 1,000 input samples, while CPU and GPU energy consumption are recorded with tegrastats~\cite{tegrastats_toolkit}. Detailed experimental settings, latency breakdowns, and further analyses are provided in the \textit{supplementary materials}.

\textbf{Autonomous Driving:} We employ ncnn~\cite{ni2017ncnn} for W8A8 support with ARM NEON intrinsic and AutoTVM~\cite{chen2018tvm} for W8A8 and W4A4 GPU kernels. As shown in Table~\ref{tab:CILRS_OpenVLA_Performance}, the \textit{CILRS} (CPU) W8A8 model achieves a 1.7$\times$ reduction in latency compared to FP32, along with a 75\% reduction in memory usage and a 30\% improvement in energy efficiency. For the \textit{CILRS} (GPU), the W4A4 model achieves up to 3.7$\times$ lower latency and 3.1$\times$ better energy efficiency.

\textbf{Robot Manipulation:} For GPU implementation of weight-only quantized \textit{OpenVLA}, we use TensorRT for the vision encoder and MLC-LLM~\cite{mlc-llm} for the LLM backbone. Compressed weights mitigate the memory-bound nature of LLM operations, significantly accelerating inference. As shown in Table~\ref{tab:OpenVLA_Performance}, the INT8 model achieves a 1.6$\times$ speedup, while the INT4 model achieves a 2.5$\times$ speedup compared to FP. Energy efficiency improves by 1.7$\times$ with INT8 and 2.5$\times$ with INT4. Additionally, the 4$\times$ memory reduction with INT4 is particularly advantageous for foundation models, which are inherently memory-intensive, making this approach suitable for edge devices with limited resources.

Our evaluations on edge devices demonstrate that the optimized models significantly improve performance and energy efficiency on both CPU and GPU platforms. These results validate that our quantization techniques enable resource-efficient robot control in computationally constrained environments.

\begin{table}[t]
\centering
\resizebox{1\columnwidth}{!}{%
\begin{tabular}{l|c|c|c|c|c|c}
\Xhline{2\arrayrulewidth}
\multirow{2}{*}{Measurement} & \multicolumn{3}{c|}{CPU (Jetson AGX Orin)} & \multicolumn{3}{c}{GPU (RTX 2080Ti)} \\ \cline{2-7}
 & FP32 & FP16 & W8A8 & FP32 & W8A8 & W4A4 \\ \midrule
Memory (MB) & 78.4 & 39.2 & 19.6 & 78.4 & 39.2 & 19.6 \\ 
Latency (ms) & 212.2 & 176.4 & 124.7 & 11.6 & 4.1 & 3.1 \\ 
SpeedUp & $1.0 \times$ & $1.2\times$ & $1.7 \times$ & $1.0 \times$ & $2.9 \times$ & $3.7 \times$ \\ 
Energy Saving & $1.0 \times$ & $1.1 \times$ & $1.3 \times$ & $1.0 \times$ & $2.5 \times$ & $3.1 \times$ \\ \Xhline{2\arrayrulewidth}
\end{tabular}
}
\vspace{-0.2 cm}
 \caption{Performance Metrics for \textit{CILRS} (14.5 GFLOPs) measured on a CPU and a GPU, demonstrating memory consumption, latency and energy saving across the different precision formats.}
    \label{tab:CILRS_OpenVLA_Performance}
\end{table}

\begin{table}[t]

    \centering
    \footnotesize
    \begin{tabular}{@{}l@{\hspace{0.1cm}}cccc@{}}
        \toprule
        \multirow{2}{*}{\centering\makecell{Weight \\ Type}} & \multicolumn{4}{c}{\textit{OpenVLA} (\( >2.0 \, \text{TFLOPs} \))} \\
        \cmidrule(lr){2-5}
         & Memory & Latency & SpeedUp & Energy Saving \\
        \midrule
        BF16 & 15.2 GB & 955.2 ms & $1.0 \times$ & $1.0 \times$ \\
        INT8 & 7.9 GB & 573.6 ms & $1.6 \times$ & $1.7 \times$ \\
        INT4 & 4.0 GB & 374.7 ms & $2.5 \times$ & $2.5 \times$ \\
        \bottomrule
    \end{tabular}
\vspace{-0.2 cm}
    \caption{Performance Metrics for \textit{OpenVLA} on GPU (NVIDIA Jetson AGX Orin): Memory, latency, and energy saving across various precision formats.}
    \label{tab:OpenVLA_Performance}
\vspace{-0.3 cm}
\end{table}

\section{Conclusion}

This paper proposes a quantization framework for IL-based models that enhances robustness against low-bit precision errors, ensuring efficiency on resource-limited hardware. Evaluations on robot manipulation and self-driving models show superior speedups and energy savings on real CPU and GPU, closely preserving full-precision accuracy and demonstrating practical deployment potential.

{
    \small
    \bibliographystyle{ieeenat_fullname}
    \bibliography{main}
}

\clearpage
\setcounter{page}{1}
\maketitlesupplementary
\setcounter{section}{0}

\section{Experiments Details}

\subsection{Benchmark Details}

\begin{figure}[h]
\centering
{\includegraphics[width=1\columnwidth]{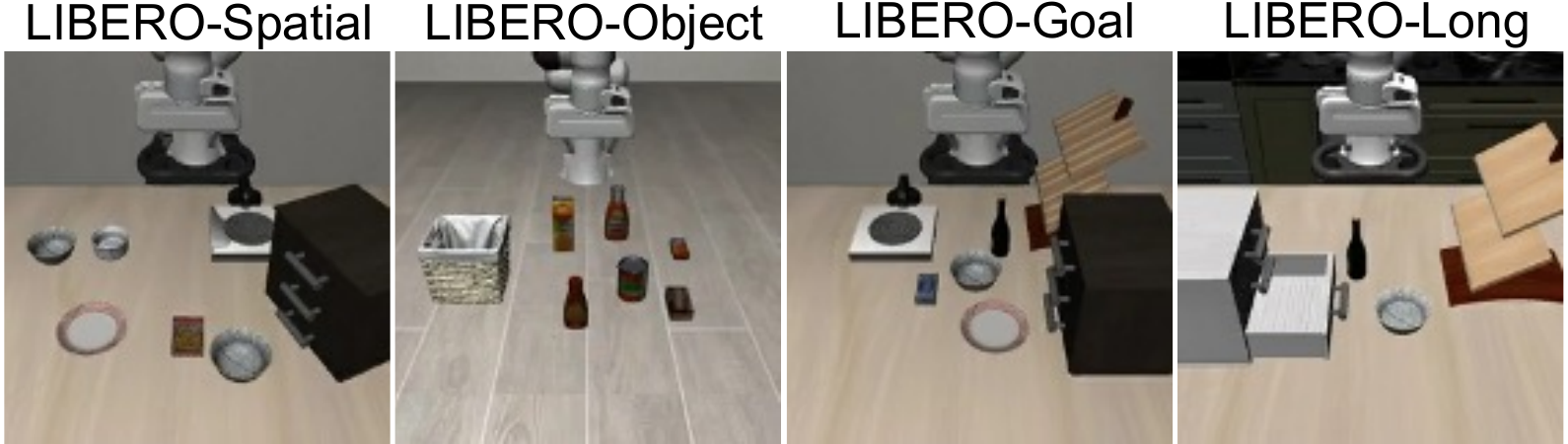}}
\caption{LIBERO benchmark.}
\label{fig:libero_bench}
\vspace{-0.3 cm}
\end{figure}

\textbf{Robot Manipulation:} We employed the LIBERO benchmark~\cite{liu2024libero} to assess the efficacy of SQIL within the realm of robot manipulation. The LIBERO benchmark includes four distinct task suites, each designed to facilitate life-long learning studies. Our research focused on implementing quantization techniques during the imitation learning process across these suites, subsequently evaluating the robustness and performance of the resulting quantized policies. Each suite comprises 10 distinct tasks accompanied by 50 human teleoperation demonstrations. LIBERO-Spatial evaluates spatial relationship understanding through varied object layouts; LIBERO-Object tests policy performance across different objects within identical layouts; LIBERO-Goal examines task-oriented behavior under consistent object and layout conditions with varied goals; and LIBERO-Long involves comprehensive long-horizon tasks that incorporate diverse objects, layouts, and objectives. To illustrate, the following are examples of tasks from each suite, with corresponding visualizations provided in Fig.~\ref{fig:libero_bench}:
\begin{itemize}
    \item \textbf{LIBERO-Spatial:} ``Pick up the black bowl \textbf{next to the ramekin} and place it on the plate."
    \item \textbf{LIBERO-Object:} ``Pick up \textbf{the butter} and place it in the basket."
    \item \textbf{LIBERO-Goal:} ``Open the middle drawer of the cabinet."
    \item \textbf{LIBERO-Long:} ``Put the black bowl in the bottom drawer of the cabinet and close it."
\end{itemize}

In our experiments, we used datasets that were specifically modified for compatibility with the OpenVLA~\cite{kim24openvla} framework, which included enhancements such as high-resolution image processing, image rotation, and the exclusion of unsuccessful demonstrations. We conducted 500 experiments for each task suite.
\begin{figure*}
    \centering
    \includegraphics[width=0.8\linewidth]{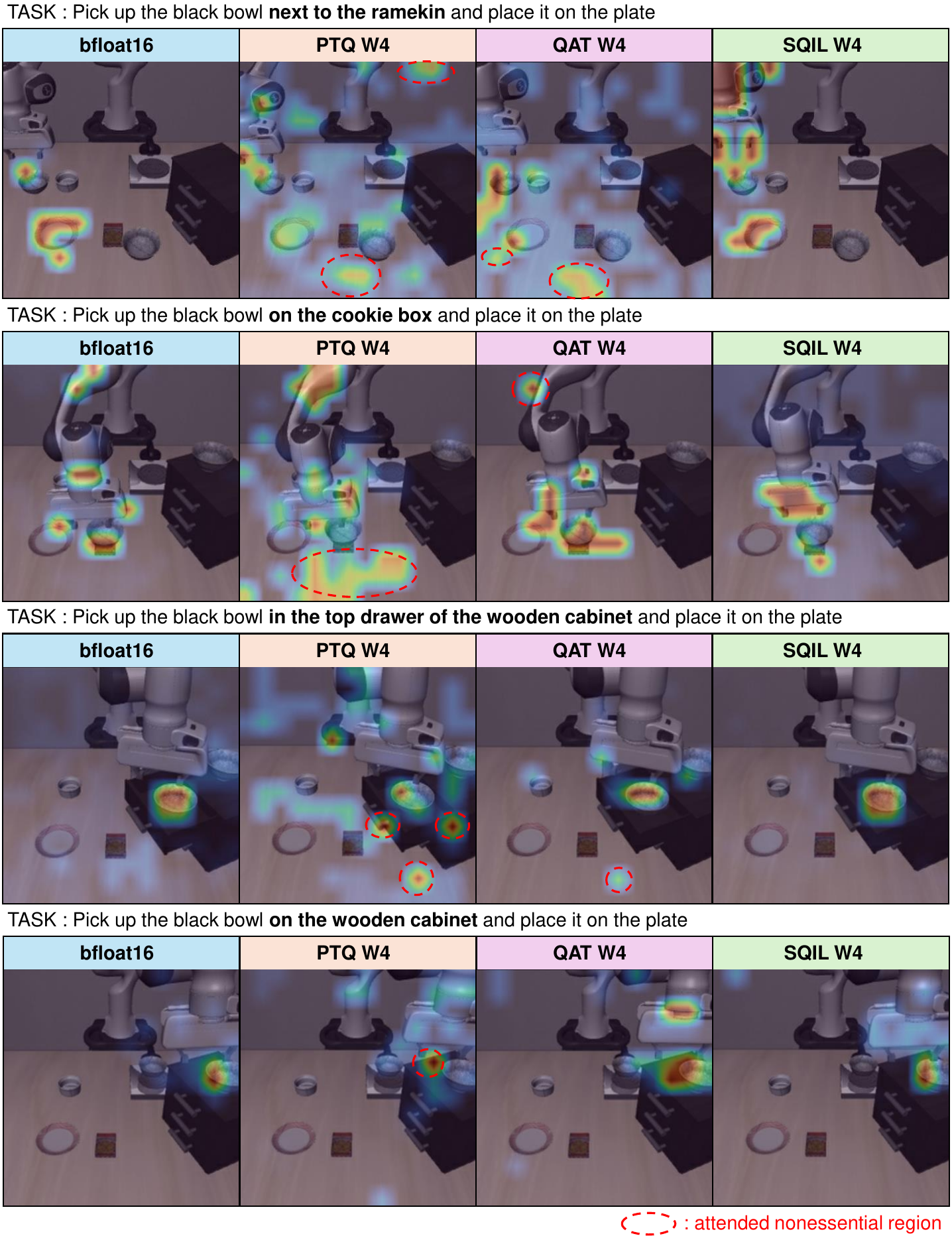}
   \caption{Comparison of attention visualization for tasks successfully completed on the LIBERO-Spatial benchmark.}

\label{fig:attention_sup}
\end{figure*}

we extended our evaluations to real-world scenarios. Using the UR5 robot, we configured an environment that mirrors the BridgeDataV2 setup, specifically adapted to resemble a toy sink setting. This setup allowed us to collect data that is comparable in complexity and variability to our simulated environments. We assembled an expert dataset by conducting 20 episodes across the following four tasks, aiming to capture diverse manipulative actions:

\begin{itemize}
    \item ``Put the eggplant into the pot."
    \item ``Put the cup into the plate."
    \item ``Stack the purple cup on the green cup."
    \item ``Lift the eggplant."
\end{itemize}
{\setlength{\parindent}{0pt}
\textbf{Autonomus Driving:}
For evaluation, we utilized the NoCrash benchmark~\cite{codevilla2019exploring}. This benchmark evaluates the generalization capabilities from Town 1, characterized by its European town setup with single-lane roads and T-junctions, to Town 2, noted for its smaller scale and distinct textural differences. The benchmark includes three traffic density levels: empty, regular, and dense. These levels define the number of pedestrians and vehicles present in each map scenario. We conducted our performance evaluations under the NoCrash-dense setting, using metrics such as the success rate proposed by NoCrash and the driving score from the CARLA LeaderBoard~\cite{leaderboard}, along with rewards derived from CARLA~\cite{dosovitskiy2017carla}. The success rate is the proportion of routes completed without collisions or blockages, while the driving score is calculated based on penalties for infractions. Our infraction analysis measures occurrences such as pedestrian and vehicle collisions and red light violations per kilometer.
}

{\setlength{\parindent}{0pt}
\textbf{Physics Simulation Tasks:}
Performance evaluation in physics simulation tasks involved measuring the average return values for each task within the DeepMind Control Suite~\cite{tassa2018deepmind}. Detailed descriptions of each task are as follows:
\begin{itemize}
    \item \textbf{Cartpole Balance}: The task requires controlling a cart to keep a pole upright by moving the cart along a track, focusing on balance and stability.
    \item \textbf{Walker Stand}: This task involves maintaining an upright posture for a bipedal walker robot without it undertaking any additional locomotion.
    \item \textbf{Hopper Stand}: A one-legged robot must remain stable and upright, testing the agent's ability to balance a dynamically unstable object.
    \item \textbf{Cheetah Run}: The goal is to maximize the forward velocity of a quadrupedal cheetah robot, emphasizing speed control and efficient limb coordination.
    \item \textbf{Finger Spin}: The agent must control a robotic finger to spin an unactuated body continuously, assessing precision and control consistency.
    \item \textbf{Humanoid Stand}: The task requires a humanoid robot to maintain a standing position, testing balance and stability in a complex robot with many degrees of freedom.
    \item \textbf{Humanoid Walk}: Extending the humanoid stand task, this requires the robot to walk at a specified speed, focusing on the coordination of bipedal locomotion.
\end{itemize}
}

\subsection{Hyperparameter Setting}
\textbf{Robot Manipulation:}
For robot manipulation tasks, we utilized channel-wise quantization for weights. Both QAT and SQIL models commence training from weights initialized using PTQ (AWQ~\cite{lin2023awq}). QLORA~\cite{dettmers2024qlora} (r=32) is employed to freeze the quantized weights, allowing updates solely to the adaptor, which comprises 110M parameters. The discrepancy metric \(D\) used is the average of the L2 distances. For SQIL, the hyper-parameter \(\beta\) is set at 2. Training proceeds wih a learning rate of 5e-4 for a total 50,000 steps.

\begin{table*}[t]
\centering
\resizebox{1\textwidth}{!}{%
\begin{tabular}{l|c|c|c|c|c|c|c|c|c|c|c}
\Xhline{2\arrayrulewidth}
\multirow{2}{*}{Method}  & \multicolumn{11}{c}{Success Rate \% $\uparrow$} \\ \cline{2-12}
 & task1 & task2 & task3 & task4 & task5 & task6 & task7 & task8 & task9 & task10& Avg.  \\ \midrule
FP & 85.3 ± 3.1 & 92.0 ± 2.8 & 83.3 ± 1.9 & 94.0 ± 4.5 & 72.7  ± 2.2 & 87.3 ± 3.1 & 91.3 ± 2.8 & 81.3 ± 1.5 & 80.0 ± 2.2 & 73.3 ± 2.9 & 84.0 \\  \midrule
 RTN&  81.3 ± 1.9 & 65.3 ± 3.1 & 74.0 ± 3.8 & 88.0 ± 2.1 & 54.7 ± 3.5 & 81.3 ± 3.0 & 92.0 ± 2.0& 76.0 ± 4.0 & 60.7 ± 2.9 & 60.0 ± 3.1& 73.3 \\ 
 AWQ&  84.7 ± 1.1 & 84.7 ± 2.1 & 72.0 ± 3.8 & 79.3 ± 1.9 & 74.0 ± 2.8 & 88.0 ± 1.5 & 92.0 ± 1.5 & \textbf{79.3 ± 1.9} & 77.3 ± 2.5 & 69.3 ± 1.9&80.1  \\ 
 QAT &  \textbf{86.0 ± 1.5} & 88.7  ± 3.1 & 77.3 ± 3.9 & 82.0 ± 2.9 & 70.7 ± 2.4 & \textbf{88.7 ± 1.8} & 90.7 ± 2.5 & 74.7 ± 1.5 & 74.0  ± 2.0 & \textbf{76.0 ± 2.1}& 80.9 \\ 
 QRD&  74.7  ± 3.3 & 78.7  ± 3.6 & 76.0  ± 4.1 & 88.0 ± 2.0 & 62.7 ± 1.3 & 75.3 ± 2.4 & 76.0 ± 4.3 & 64.7 ± 3.5 & 46.7 ± 2.1 & 52.0 ± 4.8& 69.5 \\ 
 \method&  \textbf{86.0  ± 3.1} & \textbf{92.7  ± 1.8} & \textbf{83.3  ± 3.3} & \textbf{93.3  ± 4.1} & \textbf{74.7  ± 2.2} & 86.7  ± 4.0 & \textbf{92.7  ± 2.3 } & 76.7  ± 3.4  & \textbf{79.3  ± 2.0 } & 73.3  ± 4.1&  \textbf{83.9} \\ \Xhline{2\arrayrulewidth}
\end{tabular}
}
\caption{Comparison of success rates across different tasks in LIBERO-Spatial, evaluated on \textit{OpenVLA} with INT4 quantization.}
\label{tab:libero_ablation}
\end{table*}

{\setlength{\parindent}{0pt}
\textbf{Autonomus Driving:}
In autonomous driving, tensor-wise quantization is employed for both weights and activations using LSQ~\cite{esser2020learned}. The discrepancy metric \(D\) utilized measures the average of the L2 distances of the policy network’s final logits. The quantized policy is trained over 15 epochs with a learning rate of 1e-4. For SQIL, the hyper-parameter \(\beta\) is set at 2. 
}

{\setlength{\parindent}{0pt}
\textbf{Physics Simulation Tasks:}
For physics simulation tasks, tensor-wise quantization for both weights and activations is achieved using LSQ. The discrepancy metric \(D\) used is the average of the L2 distances of the policy network’s final logits. Training uses demonstration data with a learning rate of 3e-4 and extends over 1,000,000 steps. The hyper-parameter \(\beta\) is set at 2. 
}


\section{Attention Map Analysis}
Additional visualizations are provided in Fig.~\ref{fig:attention_sup} for Sec.~\ref{sec:analysis}, illustrating a broad spectrum of tasks. The full-precision (FP) policy consistently demonstrates precise focus on relevant task objects and their specific locations, particularly where interactions with the robot arm are likely to occur. In contrast, policies quantized using PTQ (AWQ) often misdirect attention towards irrelevant areas, frequently overlooking the critical zones necessary for successful task execution. QAT represents an improvement, more accurately targeting relevant object areas, though it occasionally still attends to non-essential regions. The SQIL significantly enhances this focus, motivating a closer alignment of the quantized policy's attention with that of the FP policy on important state. This alignment contributes to actions that are more likely based on relevant and accurate situational awareness, reflecting the reasoning processes of the FP policy.

\section{Ablation Study}
\label{appendix:ablation}
\textbf{Quantization Impact Analysis.} To assess the distinct impacts of different quantization methods, we separately evaluate each approach, as detailed in Table~\ref{tab:libero_ablation}. Applying RTN quantization without any calibration for reducing quantization errors results in a significant average performance degradation of 10.7\% compared to the baseline. Using QAT, which utilizes the expert dataset as a ground truth for fine-tuning, shows performance improvements over AWQ. However, applying QRD, which aims to reduce discrepancies with the FP policy without ground truth, actually results in a decrease in performance. In contrast, \method losses brings substantial performance improvements, as \method acts as a positive guide on mission-critical state.

\textbf{ViT vs. LLM: Quantization Impact on Performance.} 
\setlength{\columnsep}{12pt}
\begin{wraptable}{r}{0.45\columnwidth}
\vspace{-0 cm}
\centering
\resizebox{0.4\columnwidth}{!}{%
\begin{tabular}{c|c|c}
\Xhline{2\arrayrulewidth}
\multicolumn{2}{c|}{\textit{OpenVLA}} & \multirow{2}{*}{Suc. Rate $\uparrow$}\\ \cline{1-2} 
ViT& LLM & \\ \midrule
FP & FP  &  74.0\%  \\ \midrule
INT4 & FP  & \textbf{73.4\%} \\
FP & INT4  & 71.3\% \\
INT4 & INT4  &70.8\% \\ \Xhline{2\arrayrulewidth}
\end{tabular}
}
\vspace{-0.3 cm}
\caption{Quantization impact of ViT and LLM components on the LIBERO for \textit{OpenVLA}.}
\vspace{-0.2 cm}
\label{tab:vit_llm}
\end{wraptable}
As demonstrated in Table~\ref{tab:vit_llm}, the quantization (PTQ) of the ViT component results in minimal performance changes, whereas quantization of the LLM component significantly impacts performance. This suggests that the reasoning capabilities of the LLM are crucial for achieving action decisions similar to those of the FP policy, indicating that \method fine-tuning effectively compensates for the losses from quantizing the LLM.

\noindent\textbf{Cost for SIS and Finetuning.}
SIS is a one-time process per existing expert dataset. We exploit spatio-temporal redundancy to reduce offline cost: perturbations are applied over a grid of $N\!\times\!N$ patches that evenly divide the image, and SIS is computed every $k$-th frame, reusing previous scores for intermediate frames. Empirically, $N\!=\!8$ and $k\!=\!4$ yielded robust results across all experiments and domains (Table~\ref{tab:timestep_frequency_sweep}). Additionally, SQIL serves as a drop-in replacement for unavoidable fine-tuning in VLA domain adaptation~\cite{black2024pi_0} due to variations (e.g., camera viewpoints, lighting, and hardware) using the same expert data and hyperparameters.

\begin{table}[h]
\begin{center}
\resizebox{0.85\linewidth}{!}{
\renewcommand{\arraystretch}{1.2}
\begin{tabular}{l|cccccc}
\Xhline{2\arrayrulewidth}
\textit{k}  & 1     & 2    & 3     & 4     & 5     & 6     \\ \hline
Suc. Rate $\uparrow$ & 49.3\%     & 49.2\%  & 49.4\%   & 49.2\%   & 48.2\%   & 47.7\%    \\ \Xhline{2\arrayrulewidth}
\end{tabular}}
\end{center}
\caption{Performance comparison of timestep frequency \( k \) for SIS calculation in \textit{OpenVLA} with SQIL on LIBERO-Long.}
\label{tab:timestep_frequency_sweep}
\end{table}
\begin{table*}[t!]
\centering
\resizebox{0.9\linewidth}{!}{%
\begin{tabular}{l|cc||ccccc||ccccc}
\Xhline{2\arrayrulewidth}
\multicolumn{3}{c||}{Discrepancy Metric $D$} & \multicolumn{5}{c||}{Saliency Weight ($\beta$)} & \multicolumn{4}{c}{Saliency Threshold ($p$)} \\
\Xhline{2\arrayrulewidth}
Setting &  L2-norm  & KL div. & 1.5 & 2 & 3 & 4 & 5 & 10\% & 15\% & 20\% & 30\% \\ 
\Xhline{2\arrayrulewidth}
Suc. Rate$\uparrow$
& \textbf{73.2\%} &  72.9\% & 49.0\% & \textbf{49.2\% }& 48.9\% & 48.2\% & 47.8\%  & 48.8\% & \textbf{49.3\%} & 49.2\% & 48.5\% \\
\Xhline{2\arrayrulewidth}
\end{tabular}
}
\caption{Comprehensive performance comparisons in the LIBERO-Spatial ($D$) and LIBERO-Long ($\beta$, \( p \)) benchmarks for \textit{OpenVLA} with \method: discrepancy metrics $D$, $\beta$ for state importance, and varying \( p \), which represents the threshold percentage of top saliency scores.}
\label{tab:hyperparameter}
\end{table*}
\begin{table}[t]
\centering
\resizebox{1.0\linewidth}{!}{
\begin{tabular}{l|cccc}
\hline
Weighting  &$\alpha_t$ & $\alpha_t$ & $\alpha_t$ & SIS  \\ 
Noise Dstribution& Gaussian  & Laplacian & Uniform & Gaussian \\ \hline
Avg. Success Rate (\%)& 73.2 ± 0.6  & 73.2 ± 0.6 & 73.0 ± 0.4 & 71.8 ± 1.0 \\
\hline
\end{tabular}}

\caption{SQIL ablations on LIBERO benchmark with \textit{OpenVLA}.}
\label{tab:reb_main}

\end{table}

\noindent\textbf{Hyperparameter Robustness.}  We examine the sensitivity of hyperparameters within the \method, as detailed in Table~\ref{tab:hyperparameter}. Opting for the L2-norm as the discrepancy metric $D$ yields slightly better results than the KL divergence. For \method, the hyperparameter $\beta$ shows stable performance within the 1.5 to 3 range. The threshold \(T\) demonstrates resilience, delivering superior performance from the top 10\% to 20\% range, and maintaining robust results even at 30\%, compared to QAT. These results confirm the robustness of our framework's hyperparameters. As shown in Table~\ref{tab:reb_main}, SQIL exhibits remarkable \textit{stability across hyperparameter variations}, enabling shared settings across domains. Changing saliency noise distribution had negligible effect, while using raw SIS values as weights degraded performance—confirming the importance of our balanced weighting approach.

\begin{figure}[t]
\centering
{\includegraphics[width=1\columnwidth]{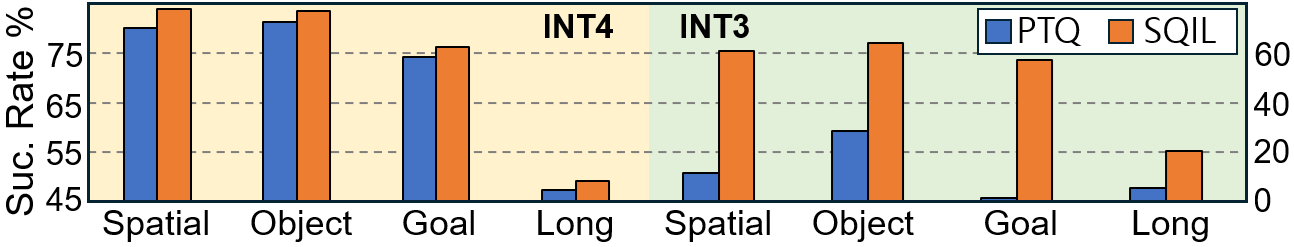}}
\caption{Performance comparison of various quantization methods across different bit precisions on \textit{OpenVLA} in the LIBERO.}
\label{fig:3bit_reb}
\end{figure}

\noindent\textbf{Aggressive Quantization.} 
We assess performance under more aggressive quantization. Fig.~\ref{fig:3bit_reb} shows that our methods enhance robustness, particularly in INT3 weight quantization of \textit{OpenVLA}.

\vspace{-0.3em}

\begin{figure*}[t]
{\includegraphics[width=\textwidth]{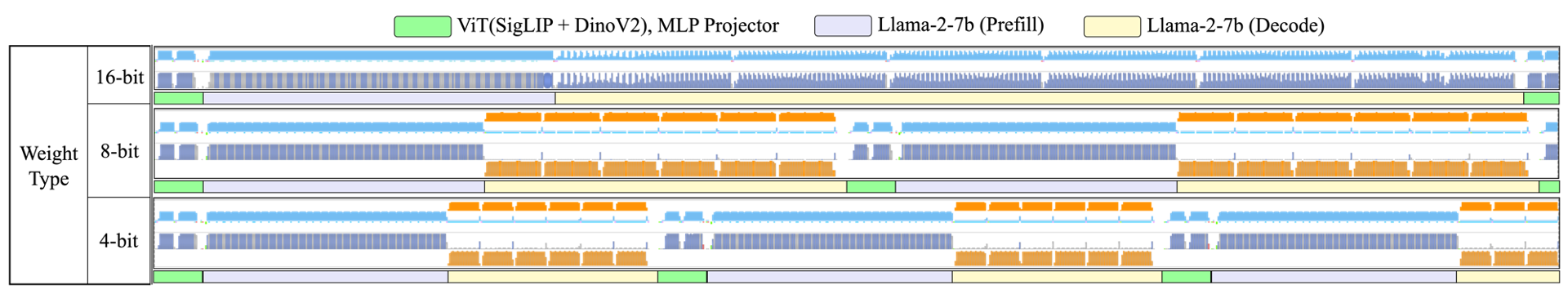}}
\caption{Hardware execution timeline comparison of 16-bit, 8-bit, and 4-bit weight datatypes for \textit{OpenVLA}, measured using \textit{NVIDIA Nsight Systems} over a total duration of 1,000 ms. The visualization highlights differences in execution patterns and latency across the three weight datatypes. Each stage is distinguished by different colors.}
\label{fig:nsys_report}
\end{figure*}

\section{Implementation Details}

\subsection{Detailed settings }

\begin{figure}[h]
{\includegraphics[width=\columnwidth]{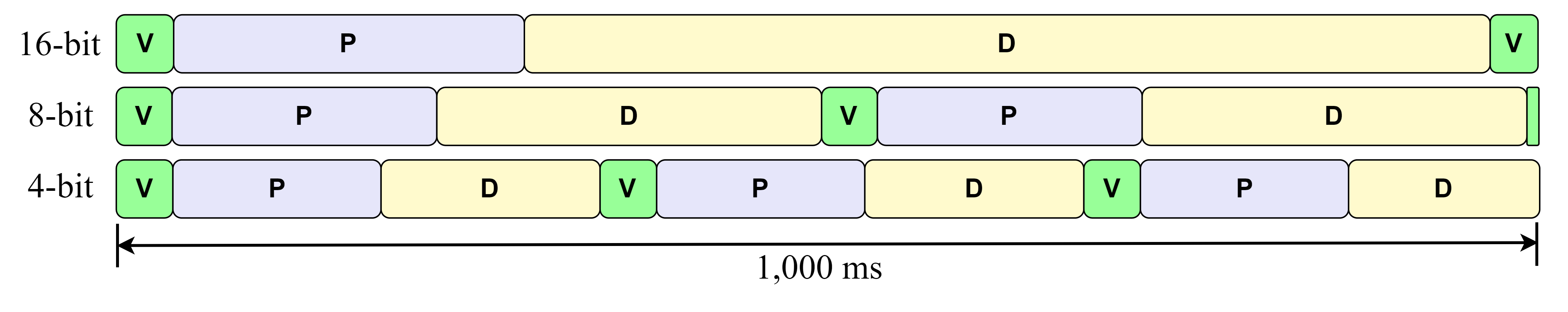}}
\caption{Timeline comparison of \textit{OpenVLA} for 16-bit, 8-bit, and 4-bit datatypes over a total duration of 1,000 ms. V represents the ViT+MLP projector, P denotes the prefill operation, and D refers to the decode operation of the backbone LLM (Llama-2-7b).}
\label{fig:timeline_openvla}
\end{figure}

\textbf{Device Settings:} For our experimental setup, we utilized NVIDIA Jetson AGX Orin 64GB and RTX 2080Ti GPU. NVIDIA Jetson AGX Orin 64GB is equipped with a 12-core Arm Cortex-A78AE CPU, an NVIDIA Ampere architecture GPU. The device runs on Ubuntu 20.04 64-bit LTS OS with GNU gcc/g++ version 9.3.0 with a 30W power mode. For each experiments, we employ the following devices:
\begin{itemize}
    \item Autonomous Driving (\textit{CILRS}, Weight-Activation Quantization):
    \begin{itemize}
        \item CPU (FP32, FP16, W8A8): NVIDIA Jetson AGX Orin 64GB (ARM Cortex-A78AE)
        \item GPU (FP32, W8A8, W4A4): NVIDIA RTX 2080 Ti
    \end{itemize}
    \item Vision-Language Action (\textit{OpenVLA}, Weight Quantization):
    \begin{itemize}
        \item GPU (16-bit, 8-bit, 4-bit): NVIDIA Jetson AGX Orin (2048-core NVIDIA Ampere GPU)
    \end{itemize}
\end{itemize} A comprehensive summary of the hardware specifications employed in our experiments is provided in Table~\ref{tab:hardware_specs_updated}.

\textbf{Energy Consumption Measurement:} To measure energy consumption in our experiments, we employ the jetson-stats library, which is specifically designed for use with NVIDIA Jetson devices. This library leverages the capabilities of the Triple Channel Voltage/Current Monitor (Texas Instrument INA3221) integrated into NVIDIA Jetson devices~\cite{tegrastats_toolkit}. The INA3221 sensor provides detailed measurements of voltage, current, and power consumption for various power rails on the device, allowing for precise monitoring and analysis of the on board power usage.
\begin{table}[t]
    \renewcommand{\arraystretch}{1.2}
    \centering
    \resizebox{\columnwidth}{!}{
    \footnotesize
    \begin{tabular}{@{}c@{\hspace{0.1cm}}cc@{}}
        \toprule
        \multirow{2}{*}{\centering\makecell{\textbf{Specification}}} & \textbf{NVIDIA Jetson AGX Orin} & \textbf{Portable Workstation} \\
        \cmidrule(lr){2-3}
         & \textbf{CPU/GPU} & \textbf{GPU} \\
        \midrule
        \makecell{\textbf{Computing} \\
        \textbf{Architecture}} 
        & \makecell{ARM Cortex-A78AE \\ 12 Cores, 2.2GHz \\ 2048-core NVIDIA Ampere \\ GPU with 64 Tensor Cores} 
        & \makecell{GeForce RTX 2080 Ti} \\
        \midrule
        \textbf{Cache (L1/L2)} 
        & \makecell{CPU: 64KB/256KB \\ GPU: 3MB/4MB} 
        & \makecell{GPU: 64KB/5.5MB} \\
        \midrule
        \textbf{Memory} 
        & 64GB LPDDR5 SDRAM 
        & 11GB GDDR6 \\
        \midrule
        \textbf{ISA} 
        & ARM v8.2-A (64 bit) / CUDA 12.0 
        & CUDA 12.0 \\
        \bottomrule
    \end{tabular}}
    \caption{Hardware Specifications of NVIDIA Jetson AGX Orin and Portable Workstation.}
    \label{tab:hardware_specs_updated}
\end{table}

\subsection{Analysis}

\begin{figure}[t]
{\includegraphics[width=\columnwidth]{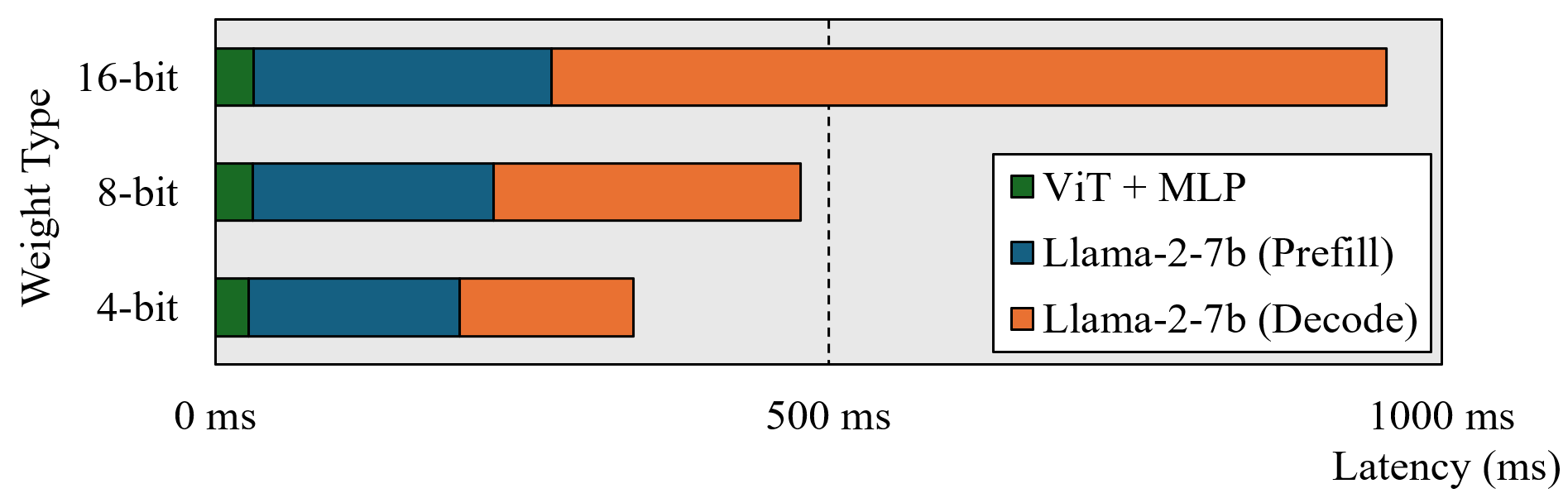}}
\caption{Latency breakdown of \textit{OpenVLA} for 16-bit, 8-bit, and 4-bit weights within a single step, including ViT+MLP, Llama-2-7b (Prefill), and Llama-2-7b (Decode) stages.}
\label{fig:latency_breakdown}
\end{figure}

We provide a detailed analysis and breakdown of the inference computation time of the \textit{OpenVLA} model. The inference process is divided into four main stages: (1) ViT (SigLIP + DINOv2), (2) MLP Projector, (3) Backbone LLM (Prefill), and (4) Backbone LLM (Decode). 

\textbf{NVIDIA Nsight Systems Hardware Execution Report:} NVIDIA Nsight Systems is a comprehensive performance analysis tool that captures and visualizes detailed hardware execution traces. In Figure~\ref{fig:nsys_report}, we present the analysis results of the hardware execution trace recorded for weight types of 16-bit, 8-bit, and 4-bit. 

\textbf{Timeline Comparison \& Breakdown:} Figure~\ref{fig:timeline_openvla} shows a timeline comparison for each weight type based on the actual time ratio, illustrating the proportion of time reduction achieved with different weight types. Figure~\ref{fig:latency_breakdown} presents the latency breakdown for each stage and analyzes how reducing the weight type affects the execution time. As shown in our experimental results, the latency of the decode operation in the backbone LLM is reduced the most, achieving up to a 2.5$\times$ speedup in overall execution time.

These experimental results confirm that the decode operation in the inference stage of the \textit{OpenVLA} model exhibits memory-bound characteristics, as demonstrated by our data. Weight quantization reduces the amount of data that needs to be transferred between memory and processing units, thereby alleviating memory bandwidth limitations and enhancing overall execution speed. By employing weight quantization techniques, we alleviate these memory-bound limitations, resulting in actual speedups in hardware execution time on Vision-Language Action model.

\label{sec:exp_details}

\end{document}